\documentclass[runningheads]{llncs}
\usepackage{graphicx}

\usepackage{tikz}
\usepackage{comment}
\usepackage{amsmath,amssymb} 
\usepackage{color}

\usepackage[breaklinks,colorlinks]{hyperref}

\begin{document}

\pagestyle{headings}
\mainmatter

\title{Facial Affect Analysis: Learning from Synthetic Data \& Multi-Task Learning Challenges}

\titlerunning{Facial Affect Analysis: LSD \& MTL Challenges}

\author{Siyang Li\inst{1}, Yifan Xu\inst{1}, Huanyu Wu\inst{1}, Dongrui Wu\inst{1}\thanks{Corresponding author}, Yingjie Yin\inst{2}, Jiajiong Cao\inst{2}, Jingting Ding\inst{2}}
\institute{Ministry of Education Key Laboratory of Image Processing and Intelligent Control, School of Artificial Intelligence and Automation, Huazhong University of Science and Technology, Wuhan, China  \and
Ant Group, Hangzhou, China\\
\email{lsyyoungll@gmail.com,  \{yfxu, m202173087, drwu\}@hust.edu.cn,   \{gaoshi.yyj, jiajiong.caojiajio, yimou.djt\}@antgroup.com}}
\authorrunning{S. Li, et al}
\maketitle

\begin{abstract}
Facial affect analysis remains a challenging task with its setting transitioned from lab-controlled to in-the-wild situations. In this paper, we present novel frameworks to handle the two challenges in the 4th Affective Behavior Analysis In-The-Wild (ABAW) competition: i) Multi-Task-Learning (MTL) Challenge and ii) Learning from Synthetic Data (LSD) Challenge. For MTL challenge, we adopt the SMM-EmotionNet with a better ensemble strategy of feature vectors. For LSD challenge, we propose respective methods to combat the problems of single labels, imbalanced distribution, fine-tuning limitations, and choice of model architectures. Experimental results on the official validation sets from the competition demonstrated that our proposed approaches outperformed baselines by a large margin. The code is available at https://github.com/sylyoung/ABAW4-HUST-ANT.

\end{abstract}

\section{Introduction} \label{introduction}
Over the years, facial affect analysis has arisen more and more attention as an important part of machine learning and computer vision. Deep neural networks have increasingly been leveraged to replace handcrafted features for automatic recognition of facial affect \cite{Li2020}. With the availability of large-scale databases, such deep models have demonstrated their ability of learning robust deep features. However, the effectiveness of data-driven approaches come with limitations. Dataset bias, imbalanced distributions,inconsistent annotations are common across most available databases, such as AffectNet \cite{Mollahosseini2019}, EmotioNet \cite{FabianBenitezQuiroz2016}.

In promoting studies of facial affect analysis, Kollias et al. \cite{Kollias2017} orgazines Affective Behavior in-the-wild (ABAW) Competition, and it has successfully been held 3 times in conjunction with the IEEE FG 2021 \cite{Kollias2020b}, ICCV 2022 \cite{Kollias2021b}, and IEEE CVPR 2022 \cite{Kollias2022}, respectively. The 4th ABAW is held in conjunction with ECCV 2022 \cite{Kollias2020a}, targeting i) Multi-Task-Learning (MTL) Challenge and ii) Learning from Synthetic Data (LSD) Challenge.  The large scale in-the-wild Aff-Wild2 database \cite{Kollias2019a,Zafeiriou2017} was used.

The tasks in facial affect analysis usually consist of the following \cite{Kollias2021,Kollias2021a,Kollias2019}: basic expressions classification, estimation of continuous affect usually including valence and arousal (VA), and detection of facial action units (AU). In the MTL challenge, all three tasks are considered.

Facial affect could be synthesized through either traditional graphic-based methods or data-driven generative models \cite{Kollias2020}, applying subtle affect change to a neutral face in the dimensional space \cite{Kollias2018,Kollias2020a}, or in categories \cite{Ding2018}. In the LSD challenge, synthetic data are given without informing the method of how the data were generated. 

The remainder of this paper is organised as follows: Section \ref{method} describes the proposed frameworks. Section \ref{experiments} presents the experimental results. Finally, Section \ref{conclusion} draws conclusions.


\section{Method} \label{method}

The proposed frameworks are discussed in this section.

\subsection{Multi-Task Learning Challenge}

\subsubsection{Model Architecture}
We used the SMM-EmotionNet~\cite{Deng2022} in multi-task learning. The feature extractor was also the inceptionV3 pre-trained on ImageNet as in~\cite{Deng2022}. We modified the usage of features in the message space of SMM-EmotionNet. Specifically, after transforming the feature vectors from the ROI embedding module, each feature was directly used to estimate dimensional emotions and emotion categories. We ensemble the predictions from different features by averaging them, i.e., continuous estimations of valence and arousal and softmax probabilities for basic emotions.

\subsubsection{Training Losses}
We used the summation of the losses in three tasks as the total loss function during training:
\begin{align}
	\mathcal{L} = \mathcal{L}^{EXPR}+\mathcal{L}^{VA}+\mathcal{L}^{AU}
\end{align}

For basic emotion classification, class-weighted cross entropy loss was used. Given a sample $x$, denote its one-hot ground-truth label and predicted softmax probabilities of each class as $y^{EXPR}$ and $\hat{y}^{EXPR} \in \mathbb{R}^{C}$, where $C$ was the number of emotion categories. The loss function for emotion classification was:
\begin{align}
		\mathcal{L}^{EXPR}(y^{EXPR}, \hat{y}^{EXPR})=-\sum _{i=1}^C w_i^{EXPR}\cdot y_i^{EXPR}\log {\hat{y}_i^{EXPR}}
\end{align}
where $w_i^{EXPR}$ was inversely proportional to the number of samples with emotion $i$ in the training set.

For valence and arousal estimation, negative concordance correlation coefficient~(CCC) of each dimension was used as the loss function. Denote the CCC between predictions and ground-truth values of valence and arousal as $CCC^V$ and $CCC^A$ respectively. The loss function was:
\begin{align}
		\mathcal{L}^{VA}= 1-CCC^V+1-CCC^A
\end{align}

For AU detection, weighted binary cross entropy loss was used. Given a sample $x$, denote its ground-truth labels and predictions of AUs as $y^{AU}$ and $\hat{y}^{AU} \in \mathbb{R}^{U}$ respectively, where $U$ was the total number of AUs to be detected. The loss function for AU detection was:
 \begin{align}
 \mathcal{L}^{AU}(y^{AU}, \hat{y}^{AU})=-\sum _{i=1}^U w_i^{AU}\cdot y_i^{AU}\log {\hat{y}_i^{AU}}+(1-y_i^{AU})\log (1-{\hat{y}_i^{AU}}),
 \end{align}
where $w_i^{AU}$ is the ratio of numbers of negative samples to positive samples of the $i$-th AU.

\subsection{Learning from Synthetic Data Challenge}

At least four challenges should be taken into consideration in training robust models using the synthetic data for expression classification:

\begin{enumerate}

\item \emph{single labels}, i.e., labels of a single class are not enough to fully reflect facial affect in expression classification.Emotions can co-occur on one face, e.,g., one can be surprised and feared at the same time. The regular 
\item \emph{imbalanced distribution} i.e., the number of training samples from each class differ by a lot. The class with the most samples (sadness) are have 10 times more samples than the class with the least samples (fear). Bias could be introduced during learning, if the issue was not carefully treated.
\item \emph{fine-tuning choice}, i.e., pretrained models are often fine-tuned on downstream task with only a fully-connected layer with limited parameters compared to the frozen feature extractor. It prevents overfitting, but also limits the models' learning abilities.
\item \emph{limitation of model architectures} i.e., specific architectures are designed to extract features with a focus on different aspects. Convolutional networks are more inclined towards local features, while Transformers have better ability to extract global features. For better analysis of facial affect, both are important aspects.
\end{enumerate}

In response to the above challenges, we propose the following framework of carefully designed strategies in accordance.

\subsubsection{Label Smoothing}
Label Smoothing \cite{Szegedy2016} was adopted in cross-entropy loss to combat the limitation of single labels. The targets become a mixture of the original ground truth and a uniform distribution. Essentially, models under label smoothing regularization during training also take into the consideration that other emotions coexist for a small likelihood other than its label class.

\subsubsection{Data Augmentations}
RandAugment \cite{Cubuk2020} was adopted to enlarge the data size of the classes with fewer samples to match the amount of class with the most samples. It consists of 14 commonly used transformations for images that was randomly combined and applied. The problem of imbalanced distribution is ameliorated through such augmentations.

\subsubsection{Fine-tuning with Deviation Module}
In order to utilize the full potential of models with millions of parameters, while preventing overfitting at the same time, we propose a more robust method for fine-tuning using pretrained models, inspired by the deviation module of DLN \cite{Zhang2021}. Specifically, a pair of siamese feature extractor with the same architecture of pretrained weights are used. One's parameters are fixed, while the other's are trainable. The actual feature vector used was their tensor discrepancy. In this way, all parameters of feature extractors are involved during fine-tuning for the downstream expression classification task.

\subsubsection{Identity-aware Pre-trained Models}
Previous works have verified that FER is closely related with face recognition. And fine-tuning models on identity-aware pre-trained weights brings significant performance gain. To this end, we adopt two different pre-training methods to increase the identity awareness of the initial models. On the one hand, we train several models on Celeb-1M dataset with identity classification task. On the other hand, we train other models on FER with large amount of generated samples. The samples are generated with different methods including face swap and expression manipulation. And the source images are with more than 100,000 identities. The final models are then developed from the pre-trained ones. 

\subsubsection{Model Backbones and Ensemble Prediction}
Four kinds of backbones, namely ViT (ViT\_B\_16), ResNet (ResNet50), EfficientNet (EfficientNet\_B0), and GoogleNet (InceptionV1) are used and separately trained on the synthetic data. The feature extractors are pretrained on ImageNet, followed by a randomly initialized fully-connected layer. At test stage of inference, an ensemble strategy is adopted that the prediction probabilities are averaged among the four models before deciding the final prediction. Specifically, the softmax scores of logits after the fully-connected layers are treated as the models' prediction probabilities.

\section{Experiments} \label{experiments}

Experimental results of the proposed approaches are shown in this section.

\subsection{Multi-Task Learning Challenge}

We used aligned faces provided by this Challenge. We first resized
the aligned faces to 299$\times$299 to fit the input size of inceptionV3. The optimizer was SGD with the momentum set to 0.9. Learning rate was set to 0.001 initially and a cosine annealing learning rate schedule was adopted. Maximum training epoch was set to 5 and the best-performing model on validation set was submitted. 

The evaluation criterion for the  Multi-Task-Learning Challenge is:

\begin{align} \label{mtll}
\mathcal{P}_{MTL} &= \mathcal{P}_{VA} + \mathcal{P}_{EXPR} + \mathcal{P}_{AU} \nonumber \\
&=  \frac{\rho_a + \rho_v}{2} + \frac{\sum_{expr} F_1^{expr}}{8} + \frac{\sum_{au} F_1^{au}}{12}
\end{align}

The performance of our model on the validation set is shown in Table~\ref{tab:mtl-result}. The evaluation metric on each task was calculated on the corresponding single-task data subset which only contained the samples with the ground-truth labels of that task.

\begin{table*}[htbp] \renewcommand{\arraystretch}{1.2}
\centering
\caption{Performance on each single-task data subset from validation set.}
\setlength{\tabcolsep}{5mm}{
\begin{tabular}{c|c|c|c}
\hline
$\mathcal{P}_{VA}$     & $\mathcal{P}_{EXPR}$     & $\mathcal{P}_{AU}$        & $\mathcal{P}_{MTL}$                  \\ \hline
0.3648        & 0.2617      & 0.4737      & 1.1002     \\ \hline
\end{tabular}}
\label{tab:mtl-result}
\end{table*}

\subsection{Learning from Synthetic Data Challenge}

\subsubsection{Details}
The amount of smoothing in Label Smoothing was empirically set to 0.2. RandAugment uses default setting of 2 transformations of magnitude 9. Solarize and Equalize are exempted from the 14 transformations because they are not suitable for applying to human faces, which would result in distorted face colors. Samples from all the other classes are combined with augmented data to match the amount of samples from the sadness class (144631). 

Images are normalized using mean and standard deviations of the synthetic training set, and resized to 224x224 before been inputted into the networks. Batch size was 64. Adam optimizer of learning rate $10^{-5}$ was used. Fine-tuning takes less than 20 epochs.

All methods were implemented using Pytorch. All computations were performed on a single GeForce RTX 3090 GPU. PyTorch was used to implement all methods. 

The official performance measure of the LSD Challenge is the average F1 Score of the 6 basic expression categories (i.e., macro F1 Score): 

\begin{align} \label{lsd}
\mathcal{P}_{LSD} &=  \frac{\sum_{expr} F_1^{expr}}{6}
\end{align}

The experimental results are shown in Table \ref{lsd_result}.

\begin{table*}[htbp]
\caption{Learning from Synthetic Data Challenge: Performance on validation set}
\label{lsd_result}
\centering
\begin{tabular}{ |c||c|c|c|c|c|c|c| }
\hline
 Method & Anger & Disgust & Fear & Happiness & Sadness & Surprise & Avg. \\
\hline
\hline
ViT & 0.4836 & 0.7110 & 0.1261 & 0.8112 & 0.5749 & 0.7140 & 0.5701 \\
 \hline
 ResNet & 0.7393 & 0.6995 & 0.1997 & 0.8219 & 0.6123 & 0.6769 & 0.6249 \\
 \hline
 EfficientNet & 0.6790 & 0.7265 & 0.2722 & 0.8038 & 0.5592 & 0.6357 & 0.6127 \\
 \hline
 GoogleNet & 0.7128 & 0.6845 & 0.2049 & 0.8030 & 0.5911 & 0.6599 & 0.6094 \\
  \hline
 Ensemble & 0.7532 & 0.7441 & 0.5631 & 0.8431 &0.6731 &0.7144  & 0.7152
 \\
 \hline
\end{tabular}
\end{table*}

\section{Conclusion}\label{conclusion}
Facial affect analysis remains a challenging task, even with the advancement of deep learning techniques. In the 4th Affective Behavior Analysis In-The-Wild
(ABAW4) Competition, we have the opportunity to contribute our efforts for better research on the problem. The carefully designed frameworks we proporsed deal with respective issues in automatic analysis of facial affect using in-the-wild dataset. Results show the effectiveness of our approaches over baseline methods.

\clearpage

\bibliographystyle{splncs04}
\bibliography{abaw}
\end{document}